# One Mastery Threshold Does Not Fit All Knowledge Tracing Models

**Xianghui MENG, Yujing ZHANG and Jionghao LIN***
*The University of Hong Kong, Hong Kong SAR, China*
*jionghao@hku.hk

**Abstract:** Tutoring systems use mastery thresholds to decide when a student can stop practicing and advance, but the same numerical threshold can mean very different things and produce very different student decisions when the underlying knowledge tracing (KT) model changes. Our study examines six KT models across four public educational datasets and evaluates 12 thresholds from 0.50 to 0.99 in terms of post-advancement performance, advancement coverage, practice burden, and disparities across prior-performance groups. We further identify the threshold that provides the trade-off between post-advancement performance, additional practice, and student advancement under 30 predefined instructional settings. Bayesian Knowledge Tracing (BKT) is relatively insensitive to threshold changes, whereas neural models become substantially more selective as thresholds increase. This difference partly reflects the models' outputs: BKT estimates the probability of a latent mastery state, whereas the neural models estimate the likelihood of a correct next response, so the same numerical cutoff does not represent the same level of mastery. The threshold that best balanced these instructional outcomes varied substantially across models and settings, showing that no single cutoff was consistently appropriate. In half of the tested settings, neural models and BKT differed by more than 0.10 in their selected thresholds, but this difference became smaller when greater priority was placed on reducing extra practice and allowing more students to advance. Stricter thresholds also did not reliably reduce performance gaps and could disproportionately restrict advancement, with stronger-prior students advancing up to 3.26 times as often as weaker-prior students. These findings show that mastery thresholds should be recalibrated when the KT model or instructional priorities change and evaluated through their consequences for student performance, practice, advancement, and access.

## 1. Introduction

Mastery learning is built on a simple instructional principle: students should demonstrate sufficient success on a skill before moving to the next one (Bloom, 1984; Corbett & Anderson, 1994). Intelligent tutoring systems operationalize this principle through a mastery threshold, a numerical cutoff applied to a knowledge tracing (KT) model's output to determine whether a student can stop practicing and advance. The familiar threshold of ($\tau$=0.95) was developed in the context of Bayesian Knowledge Tracing (BKT), where the model estimates the probability of a latent mastery state (Corbett & Anderson, 1994; Anderson et al., 1995). Neural knowledge-tracing models, such as Deep Knowledge Tracing (DKT; Piech et al., 2015), instead estimate the likelihood of a correct next response. The same numerical threshold therefore does not necessarily represent the same level of mastery across KT architectures.

This difference has direct instructional consequences because mastery thresholds determine how long students continue practicing and whether they gain access to later material. Prior studies have shown that mastery criteria can influence subsequent performance and retention (Matayoshi et al., 2025; Zhang et al., 2025), but threshold selection has largely been examined within individual tutoring systems or model families. It therefore remains unclear whether a threshold selected for one KT architecture can be applied to another without changing the resulting instructional decisions. Rather than interpreting a cutoff as a universal probability that a student knows a skill, its appropriateness should be evaluated through observable consequences, including post-advancement performance, advancement coverage, practice burden, and disparities across prior-performance groups. We examine these consequences across six KT models and four public educational datasets. We aim to answer three **R**esearch **Q**uestions (**RQ**s):

- **RQ1**. How does mastery threshold choice affect post-advancement performance, advancement coverage, and practice burden across KT model architectures?
- **RQ2**. How do stricter mastery thresholds affect post-advancement performance gaps and advancement disparities across prior-performance groups?
- **RQ3**. How do KT model architecture and instructional priorities determine the threshold that best balances post-advancement performance, practice burden, and student advancement?

This study makes three contributions. First, it demonstrates that the same numerical threshold can produce substantially different advancement decisions across KT architectures. For example, on ASSISTments, a threshold of 0.80 results in advancement coverage of 0.889 under BKT but only 0.113 under DKT. Second, it shows that stricter thresholds do not reliably reduce post-advancement performance gaps and can disproportionately restrict advancement for students with weaker prior performance, with stronger-prior students advancing up to 3.26 times as often as weaker-prior students. Third, rather than assuming one universally appropriate threshold, we identify thresholds under 30 predefined instructional settings that place different priorities on post-advancement performance, additional practice, and student advancement. The selected thresholds vary substantially across models and instructional priorities, showing that mastery thresholds should be recalibrated when either the KT model or the instructional objective changes.

## 2. Related Work

Mastery learning requires students to demonstrate sufficient proficiency before advancing (Bloom, 1984), and early intelligent tutoring systems commonly implemented this principle using Bayesian Knowledge Tracing (BKT), often with a threshold of $\tau$=0.95 on the estimated latent mastery state (Corbett & Anderson, 1994; Anderson et al., 1995). Prior studies show that mastery criteria can affect both practice and later performance. Pelánek and Řihák (2017) compared mastery criteria in terms of prediction and practice efficiency. In a randomized ALEKS study, Matayoshi et al. (2025) found that a stricter mastery rule modestly improved retention, although the effect varied across students and subject areas. Zhang et al. (2025) similarly found that higher BKT mastery estimates were associated with stronger subsequent performance. These studies demonstrate that mastery thresholds have instructional consequences, but they examine thresholds within a particular system or model family.

Whether the same numerical threshold can be transferred across KT architectures remains less clear. BKT estimates a latent mastery state, while Performance Factors Analysis (PFA) accumulates evidence from prior successes and failures, DKT produces recurrent next-response predictions, and SAKT, AKT, and SAINT use attention-based representations (Pavlik et al., 2009; Piech et al., 2015; Pandey & Karypis, 2019; Choi, Lee, Cho, et al., 2020; Ghosh et al., 2020). Consequently, similar numerical scores across these models do not necessarily represent the same level of student mastery, even when models achieve similar predictive performance (Gervet et al., 2020). Applying the same mastery threshold across architectures may therefore produce different advancement decisions.

Related fairness research has largely focused on improving the estimation produced by student models. For example, Barrett et al. (2024) incorporated response time into BKT to reduce gender- and income-related bias, while Zhang et al. (2024) used causal methods to reduce the influence of sensitive attributes in cognitive diagnosis. However, less attention has been paid to the decision layer, where model outputs are converted into decisions about whether students continue practicing or advance. We address this gap by examining how mastery thresholds affect post-advancement performance, advancement coverage, practice burden, and disparities across prior-performance groups across different KT architectures. We further examine how the threshold that best balances these outcomes changes under 30 predefined instructional settings with different priorities for student performance, additional practice, and advancement.

## 3. Method

Figure 1 summarizes the architecture-aware threshold calibration workflow. We evaluate six knowledge tracing models, grouped into classical and neural architectures. Because these models produce outputs with different statistical meanings, we treat each output as a model-specific operational advancement signal rather than assuming that all scores represent the same probability of knowledge. We apply 12 thresholds from 0.50 to 0.99 and evaluate post-mastery hit rate, advancement coverage, attempts to mastery, and equity outcomes. We then examine utility-maximizing thresholds across a frozen 30-setting penalty surface.

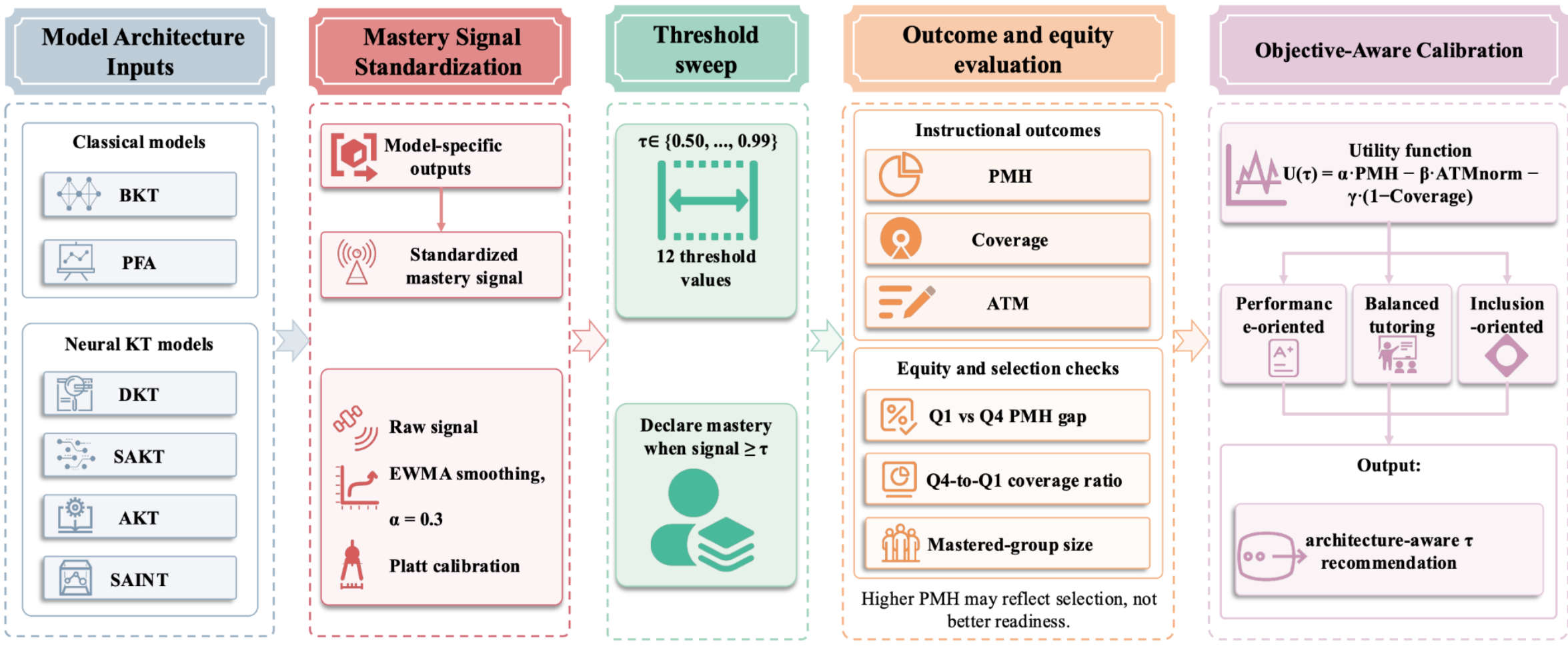

*Figure 1.* Architecture-aware threshold calibration workflow for knowledge tracing models. The workflow applies model-specific advancement signals to 12 candidate thresholds, evaluates post-advancement performance, coverage, practice burden, and equity, and examines utility-maximizing thresholds across a 30-setting instructional-penalty surface.

### *3.1 Datasets*

Table 1. *Dataset Statistics After Preprocessing and Event Capping*

| Dataset | Student | Knowledge Components | Interaction Records | Avg. Interactions per Student | Correctness Rate (%) |
|---|---|---|---|---|---|
| ASSISTments 2009 | 2,254 | 138 | 200,000 | 88.7 | 62.4 |
| EdNet KT1 | 352 | 187 | 200,000 | 568.2 | 55.1 |
| DBE-KT22 | 543 | 13 | 19,400 | 35.7 | 71.8 |
| Junyi Academy | 816 | 1,278 | 189,000 | 231.6 | 69.8 |

*Note. Knowledge components are the distinct skills tracked after preprocessing. Correctness rate is the proportion of correct interactions. Junyi Academy contains an average of 10.8 interactions per student-skill pair.*

We use four public datasets: ASSISTments 2009, which contains U.S. middle-school mathematics practice (Feng et al., 2009); EdNet KT1, which records Korean test-preparation activity (Choi, Lee, Shin, et al., 2020); DBE-KT22, which contains student interactions from a university database course in Australia (Abdelrahman et al., 2022); and Junyi Academy, which contains Taiwanese K-12 mathematics practice (Chang et al., 2015). Table 1 summarizes their properties. Interactions are chronologically ordered so that no model uses future responses when estimating an earlier state. Each dataset is capped at 200,000 interactions after preprocessing; DBE-KT22 and Junyi Academy contain fewer records and therefore use all available interactions. Junyi Academy has relatively sparse skill-level histories, with 10.8 interactions per student-skill pair, while DBE-KT22 contains only 13 knowledge components and has a correctness rate of 71.8%. These characteristics can make high thresholds easier to reach and limit generalization to systems with more skills, lower correctness rates, or sparser learner histories. All analyses used publicly released, de-identified datasets and involved no recruitment, intervention, or direct interaction with human participants. The datasets were accessed under their publicly stated repository terms.

### *3.2 Knowledge Tracing Models*

All six models are trained and evaluated under a user-level 70/10/20 *train, validation, and test* split, keeping every learner in only one partition to prevent leakage. Bayesian Knowledge Tracing (BKT) models each skill as a latent binary state updated through prior-knowledge, learning, guess, and slip parameters (Corbett & Anderson, 1994). Performance Factors Analysis (PFA) models the log-odds of a correct response from cumulative successes and failures for each skill (Pavlik et al., 2009). Deep Knowledge Tracing (DKT) maps response histories to next-response predictions using a recurrent neural network (Piech et al., 2015). SAKT, AKT, and SAINT use attention-based sequence architectures (Pandey & Karypis, 2019; Ghosh et al., 2020; Choi, Lee, Cho, et al., 2020). Neural models use early stopping with a patience of 10 epochs based on validation loss. Main analyses use fixed random seeds, common user-level splits, and shared model configurations across datasets. Supplementary multi-seed analyses assess the stability of predictive performance and threshold-sensitivity patterns. We report the held-out area under the receiver operating characteristic curve and log-loss as predictive references. Higher AUC indicates stronger discrimination, while lower log-loss indicates better probabilistic prediction. These metrics do not determine whether a threshold is instructionally appropriate because models with similar predictive quality can still produce different score distributions and advancement decisions (Pelánek, 2015). For BKT, the advancement signal is the posterior latent-state probability (P(Know)). For predictive evaluation only, this quantity is converted to the implied probability of a correct next response:

$$P(correct) = P(Know) \times (1 - slip) + (1 - P(Know)) \times guess \tag{1}$$

### *3.3 Model-Specific Advancement Signals*

The six KT models do not produce a common probability of mastery. For Bayesian Knowledge Tracing (BKT), the advancement signal is the posterior probability of a latent knowledge state, P(Know). Performance Factors Analysis (PFA) and the neural models instead produce scores related to predicted next-response correctness. BKT and PFA use their direct outputs in the main analysis. For DKT, SAKT, AKT, and SAINT, we apply an exponentially weighted moving average (EWMA) with a smoothing coefficient of 0.3 to reduce variation caused by individual responses. At each opportunity, the smoothed signal assigns a weight of 0.3 to the current predicted response probability and a weight of 0.7 to the previous smoothed signal.

Robustness analyses additionally compare raw, EWMA-smoothed, and Platt-calibrated neural signals (Guo et al., 2017; Platt, 1999). Platt calibration can improve correspondence between predicted and observed correctness, but it does not transform a neural model output into a probability that a student has mastered the skill. We therefore treat all thresholded outputs as model-specific advancement signals rather than assuming that the same numerical value has the same interpretation across KT models.

### *3.4 Threshold Policies and Outcome Metrics*

For each model, a student-skill sequence is considered ready to advance when its model-specific advancement signal first reaches or exceeds the selected threshold. We evaluate 12 thresholds: 0.50, 0.55, 0.60, 0.65, 0.70, 0.75, 0.80, 0.85, 0.90, 0.95, 0.98, and 0.99. Robustness analyses additionally require the threshold to be maintained for two or three consecutive opportunities before advancement.

We evaluate each threshold using three instructional outcomes. Post-mastery hit rate (PMH5) is the fraction of correct responses among up to the first five available responses after threshold crossing. It measures observed post-advancement performance and is not interpreted as direct evidence of latent mastery. Advancement coverage is the proportion of evaluated student-skill sequences that reach the threshold. Attempts to mastery, abbreviated as ATM, is the number of opportunities up to and including the first threshold crossing, averaged among sequences that advance. These outcomes must be interpreted together because a stricter threshold can raise PMH5 by selecting a smaller and stronger group rather than by improving readiness among all learners. We flag estimates based on fewer than 30 advancing sequences as unstable.

For the equity analysis, learners are divided into prior-performance quartiles using training-split accuracy, with Q1 representing the weakest prior records and Q4 the strongest. At each threshold, we compare group-specific advancement coverage and PMH5, the Q4-minus-Q1 PMH5 gap, and the Q4-to-Q1 advancement-coverage ratio. We also conduct selection-sensitivity analyses using inverse-probability-weighted PMH5 and Manski-style bounds. As an actionable equity sensitivity analysis, we also examine whether candidate thresholds satisfy illustrative deployment guardrails. The tested guardrails require minimum Q1 advancement coverage of 0.25, 0.40, or 0.50 and a maximum Q4-to-Q1 advancement-coverage ratio of 1.5 or 2.0. These post-hoc guardrails identify thresholds that limit exclusion and between-group disparity. They are deployment constraints rather than causal interventions. All confidence intervals use learner-level bootstrap resampling. This procedure resamples complete learners and retains their associated skill sequences, accounting for dependence among responses from the same learner.

*3.5 Decision Utility and Cross-Architecture Transfer*

To make the trade-off among post-advancement performance, practice burden, and advancement coverage explicit, we define the utility of threshold τ as:

$$U(\tau) = PMH_5(\tau) - a \cdot ATM_n(\tau) - c \cdot [1 - Coverage(\tau)] \quad (2)$$

Here, $\tau$ is the candidate threshold, and $ATM_n$ is attempts to mastery min-max normalized across the candidate thresholds within the same model-dataset combination. The coefficient on $PMH_5$ is fixed at 1. The coefficient a represents the importance placed on limiting additional practice, while $c$ represents the importance placed on allowing more student-skill sequences to reach advancement. We evaluate five values of a (0, 0.02, 0.05, 0.10, and 0.20) and six values of $c$ (0.02, 0.05, 0.10, 0.20, 0.30, and 0.50), producing 30 predefined instructional settings for each model and dataset. For each setting, we select the threshold with the highest utility. The selected threshold therefore reflects the best trade-off for a particular model, dataset, and set of instructional priorities rather than a universal mastery threshold.

As a secondary analysis, we examine whether a threshold selected for one KT model can be transferred to another without recalibration. Threshold-transfer loss is the difference between the target model's highest achievable utility and the utility obtained when it uses a threshold selected for another model. A larger loss indicates that transferring the threshold leads to a less favorable trade-off among post-advancement performance, practice burden, and advancement coverage. Because this loss depends on the instructional priorities, it is interpreted as policy-specific rather than as a fixed property of a model pair. The transfer analysis uses one predefined setting, $a$=0 and $c$=0.05, excludes transfers from a model to itself, and omits cases in which utility cannot be calculated.

## 4. Results

*4.1 Model Prediction Quality*

Table 2 reports held-out prediction quality as context for the threshold analyses. Neural KT models generally achieved higher AUC than Bayesian Knowledge Tracing (BKT), particularly on ASSISTments, DBE-KT22, and Junyi Academy, while Performance Factors Analysis (PFA) generally fell between BKT and the neural models. However, stronger next-response prediction did not necessarily lead to more appropriate advancement decisions. Models with similar or higher predictive performance could still produce substantially different score distributions and advancement coverage under the same threshold. The remaining analyses therefore focus on the instructional consequences of threshold choice.

Table 2. *Held-Out Prediction Quality for the Six KT Models*

| Model | ASSISTments | | EdNet KT1 | | DBE-KT22 | | Junyi Academy | |
|---|---|---|---|---|---|---|---|---|
| | AUC | LL | AUC | LL | AUC | LL | AUC | LL |
| BKT | 0.567 | 0.595 | 0.571 | 0.606 | 0.514 | 0.472 | 0.508 | 0.589 |
| PFA | 0.709 | 0.590 | 0.623 | 0.610 | 0.711 | 0.530 | 0.672 | 0.660 |
| DKT | 0.750 | 0.550 | 0.628 | 0.580 | 0.772 | 0.440 | 0.701 | 0.570 |
| SAKT | 0.734 | 0.560 | 0.622 | 0.580 | 0.770 | 0.440 | 0.686 | 0.580 |
| AKT | 0.730 | 0.560 | 0.597 | 0.610 | 0.766 | 0.440 | 0.684 | 0.580 |
| SAINT | 0.713 | 0.570 | 0.596 | 0.620 | 0.765 | 0.440 | 0.691 | 0.590 |

*Note. AUC = area under the ROC curve; LL = log-loss. Higher AUC and lower LL indicate better prediction quality.*

### *4.2 The Same Threshold Produces Different Advancement Decisions Across KT Models*

The effect of increasing the mastery threshold differed substantially across KT models (Table 3). BKT was relatively insensitive to threshold changes. As the threshold increased from 0.50 to 0.80, its post-advancement performance changed by no more than 0.005 across the four datasets, while advancement coverage remained above 0.57 in every dataset and reached 0.889 on ASSISTments. In practical terms, increasing the BKT threshold over this range changed relatively few advancement decisions.

Table 3. *Threshold Sensitivity Under the Threshold-Only Mastery Rule*

*(a) Change in $PMH_5$ from $\tau$ = 0.50 to $\tau$ = 0.80*

| Model | ASSISTments | EdNet KT1 | DBE-KT22 | Junyi Academy |
|---|---|---|---|---|
| BKT | +0.004 | −0.005 | −0.001 | +0.003 |
| PFA | +0.134 | +0.175 | +0.076 | +0.066 |
| DKT | +0.131 | N/A | −0.001 | +0.104 |
| SAKT | +0.128 | +0.162 | −0.001 | +0.109 |
| AKT | +0.115 | N/A | 0.000 | +0.104 |
| SAINT | +0.136 | +0.105 | 0.000 | +0.078 |

*(b) Advancement Coverage at t = 0.80*

| Model | ASSISTments | EdNet KT1 | DBE-KT22 | Junyi Academy |
|---|---|---|---|---|
| BKT | 0.889 | 0.573 | 0.667 | 0.986 |
| PFA | 0.332 | 0.045 | 0.560 | 0.570 |
| DKT | 0.113 | N/A | 1.000 | 0.047 |
| SAKT | 0.114 | 0.028 | 1.000 | 0.049 |
| AKT | 0.127 | N/A | 1.000 | 0.048 |
| SAINT | 0.069 | 0.003 | 1.000 | 0.061 |

*Note. PMH5 is the fraction of correct responses in up to the first five available items after threshold crossing. Advancement coverage is calculated over evaluated student-skill sequences. N/A indicates advancement coverage below 0.01, for which PMH5 is not reported. On ASSISTments at $\tau$ = 0.80, attempts to advancement were 1.419 for BKT and 2.363 for DKT.*

The neural models showed a different pattern. On ASSISTments, for example, DKT's post-advancement performance increased by 0.131 as the threshold rose from 0.50 to 0.80, but advancement coverage fell to 0.113. Thus, fewer than 12% of evaluated student-skill sequences reached advancement at the 0.80 threshold. The practice burden among advancing sequences was also higher under DKT, with 2.363 attempts to advancement compared with 1.419 under BKT. EdNet KT1 showed a stronger reduction in advancement coverage, with DKT and AKT falling below 0.01 before reaching the 0.80 threshold.

DBE-KT22 was the main exception. Neural models retained high advancement coverage at a threshold of 0.80, reflecting dataset characteristics such as its small number of knowledge components and relatively high correctness rate. This result indicates that threshold sensitivity depends not only on model architecture but also on the characteristics of the student interaction data. Overall, the same numerical threshold did not correspond to the same level of instructional strictness across KT models. A threshold of 0.80 allowed broad advancement under BKT but became highly restrictive for several neural models.

Robustness analyses across post-advancement evaluation horizons produced the same overall pattern. We recomputed post-advancement performance using the first 5, 10, and 20 available responses after threshold crossing. On ASSISTments, the neural-versus-BKT sensitivity gap remained nearly unchanged, from 0.124 at H=5 to 0.125 at H=20. On EdNet KT1, the corresponding gap changed from 0.232 to 0.215, while BKT sensitivity remained approximately 0.003 across all three horizons. Thus, the architecture-dependent pattern was not specific to the five-response evaluation window.

### *4.3 Stricter Thresholds Do Not Reliably Reduce Prior-Performance Gaps*

Across all six models and four datasets, raising the threshold does not reliably reduce the PMH5 gap between the lowest and highest prior-performance groups. In some model-dataset settings, the Q4-to-Q1 PMH5 gap changed little. In others, the gap appeared to narrow partly because fewer Q1 student-skill sequences reached the higher threshold, leaving a smaller and more selected group for post-advancement evaluation. Thus, a smaller performance gap at a stricter threshold does not necessarily indicate improved outcomes for weaker-prior students. Table 4 summarizes these prior-performance differences in both post-advancement performance and advancement coverage

Table 4. *Prior-Performance Equity Summary Across Six KT Models and Four Datasets*

| Equity Outcome | Finding | Interpretation |
|---|---|---|
| **PMH gaps** | Stricter thresholds do not reliably reduce Q4-Q1 PMH gaps | Higher thresholds do not consistently improve post-mastery equity |
| **Coverage disparity** | Maximum Q4-to-Q1 advancement-coverage ratio = 3.26 | Stronger-prior students are much more likely to reach mastery at high thresholds |
| **Small mastered groups** | At $\tau \geq 0.90$, 57% of configurations have < 30 mastered students | High-threshold PMH estimates are frequently unstable |
| **Deployment note** | Report PMH with coverage, group ratios, and mastered-group size | Strict thresholds alone are not an equity intervention |

Note. Q1 = lowest prior-performance quartile; Q4 = highest. Prior-performance groups defined by training-split correctness rate.

The clearest effect is on advancement access. Across the evaluated settings, the Q4-to-Q1 advancement-coverage ratio reached as high as 3.26, meaning that student-skill sequences from the strongest-prior group were, in some settings, more than three times as likely to reach advancement as those from the weakest-prior group. This pattern reflects how the threshold rule distributes advancement opportunities across prior-performance groups and should not be interpreted as a causal effect on individual students.

We therefore examined illustrative deployment guardrails that constrain both minimum Q1 advancement coverage and the Q4-to-Q1 coverage ratio. At least one candidate threshold satisfying the tested constraints was available in 81.25% of evaluated model-dataset settings. The median reduction in utility associated with applying these guardrails was 0.072, with a maximum of 0.221. These results suggest that threshold selection can explicitly consider advancement disparities, although the guardrails do not establish causal improvements in learning or fairness.

*4.4 Thresholds Depend on Both the KT Model and Instructional Priorities*

The threshold that best balanced post-advancement performance, practice burden, and advancement coverage varied substantially across the 30 predefined instructional settings. Across these settings, the mean selected threshold ranged from 0.500 to 0.954 for neural models and from 0.500 to 0.890 for Bayesian Knowledge Tracing (BKT). In 15 of the 30 settings, the mean thresholds selected for neural models and BKT differed by more than 0.10. These results indicate that the same threshold is not consistently appropriate across model architectures.

The selected threshold also depended on the instructional priorities. When greater importance was placed on limiting additional practice and allowing more student-skill sequences to advance, both BKT and neural models selected lower thresholds. In nine settings, both model families selected the lowest candidate threshold of 0.50. In contrast, neural models selected mean thresholds of at least 0.90 in seven settings and no greater than 0.55 in twelve settings. Thus, neither a high threshold for neural models nor a low threshold for BKT was consistently preferable. Instead, threshold selection depended jointly on the model, dataset, advancement signal, and instructional priorities.

As a secondary analysis, we examine the utility loss that occurs when a threshold selected for one model is transferred to another model without recalibration. This analysis uses the frozen setting $a = 0$ and $c = 0.05$. Classical-to-neural transfer has a mean utility loss of 0.0597 across 26 valid transfers, with a median of 0.0081 and a maximum of 0.2212. Neural-to-classical transfer has a mean loss of 0.0175 across 32 valid transfers, with a median of 0.0070 and a maximum of 0.0984. Within-neural transfer has a mean loss of 0.0102 across 42 valid transfers, while within-classical transfer has a mean loss of 0.0577 across 8 valid transfers. Because these values depend on the selected penalty setting and several transfer cells are undefined, they should be interpreted as policy-conditional sensitivity results rather than as a universal ranking of transfer directions.

## 6. Discussion and Conclusion

This study examined how mastery thresholds translate knowledge tracing (KT) model outputs into advancement decisions across six models and four educational datasets. *First*, the same numerical threshold can produce substantially different advancement patterns across KT models. *Second*, stricter thresholds do not consistently reduce post-advancement performance gaps and may disproportionately restrict advancement for students with weaker prior performance. *Third*, the threshold that best balances post-advancement performance, practice burden, and advancement coverage changes with both the KT model and the instructional priorities. These findings indicate that a mastery threshold should be treated as an instructional decision rule whose consequences must be evaluated, rather than as a fixed technical setting that can be transferred across models.

A key implication is that the numerical value of a mastery threshold has meaning only in relation to the model that produces the advancement signal. BKT thresholds a posterior probability of a latent mastery state, whereas neural KT models use signals related to predicted next-response correctness. A threshold such as 0.80 therefore should not be assumed to represent the same level of student mastery across architectures. This distinction is especially important when communicating KT outputs to teachers or using them to configure tutoring systems. Rather than interpreting a cutoff as a literal probability that a student knows a skill, systems should report its observable consequences, such as post-advancement performance, advancement coverage, and additional practice. This interpretation extends prior work showing that mastery criteria influence practice, retention, and subsequent performance (Matayoshi et al., 2025; Zhang et al., 2025) by showing that these consequences also depend on the architecture generating the thresholded score.

The findings also suggest that stricter mastery rules are not automatically better instructional policies. Higher thresholds may produce stronger post-advancement performance partly because fewer and stronger-performing student-skill sequences remain eligible to advance. They can therefore improve the observed performance of the advancing group while simultaneously reducing access to advancement, particularly for students with weaker prior performance. This result resonates with prior research showing that mastery rules and adaptive decisions can shape students' access to later learning opportunities (Doroudi, 2020) and complements fairness work that has focused primarily on improving the estimation produced by student models (Barrett et al., 2024; Zhang et al., 2024). The present study highlights an additional decision-layer concern: even a well-performing student model can produce unequal advancement opportunities when its outputs are combined with a particular threshold. In practice, threshold selection should therefore consider not only post-advancement performance but also practice burden, advancement coverage, and disparities across prior-performance groups. When a platform changes its KT model or instructional priorities, the threshold should be recalibrated rather than transferred automatically.

Several limitations define the scope of these conclusions. The study uses observational interaction logs rather than randomly assigning students to different mastery thresholds, so differences in post-advancement performance should not be interpreted as causal effects of stricter thresholds. The primary outcome measures performance over a limited number of responses after threshold crossing, and the prior-performance analysis does not examine protected demographic groups. The tested guardrails are illustrative deployment constraints rather than evidence that a threshold improves fairness or learning. In addition, the evaluated models do not include KT systems integrated with large language models or personalized thresholds for individual students. Selection also remains important because students who reach stricter thresholds may systematically differ from those who do not. Future work should therefore test threshold policies through longitudinal and randomized studies, examine retention and transfer to related skills, evaluate demographic disparities, and investigate whether personalized or adaptive thresholds can better balance student performance, practice burden, and advancement access.

## Acknowledgements

This work was supported by the Research Grants Council (Hong Kong) under Early Career Scheme, Grant/Award Number: #27608026. The opinions, findings, and conclusions expressed in this paper are solely those of the authors.